\documentclass[conference,10pt]{IEEEtran}
\IEEEoverridecommandlockouts
\usepackage{amsmath,amssymb,amsfonts}
\usepackage{algorithmic}
\usepackage{graphicx}
\usepackage{textcomp}
\usepackage{xcolor}
\usepackage{glossaries}
\usepackage{cite}
\usepackage{comment}
\usepackage{siunitx}
\usepackage{subcaption}
\usepackage{multirow}

\usepackage{hyperref}
\newcolumntype{L}[1]{>{\raggedright\let\newline\\\arraybackslash\hspace{0pt}}m{#1}}
\newcolumntype{C}[1]{>{\centering\let\newline\\\arraybackslash\hspace{0pt}}m{#1}}
\newcolumntype{R}[1]{>{\raggedleft\let\newline\\\arraybackslash\hspace{0pt}}m{#1}}

\usepackage{booktabs}

\usepackage[capitalise]{cleveref}

\usepackage[a4paper, total={184mm,239mm}]{geometry}
\def\BibTeX{{\rm B\kern-.05em{\sc i\kern-.025em b}\kern-.08em
    T\kern-.1667em\lower.7ex\hbox{E}\kern-.125emX}}

\usepackage{eso-pic}
\usepackage{url}
\AddToShipoutPictureBG*{
  \AtPageUpperLeft{%
    \put(0,-40){\raisebox{15pt}{\makebox[\paperwidth]{\begin{minipage}{21cm}\centering
      \textcolor{gray}{This article has been accepted for publication in the proceedings of the \\2024 IEEE Sensors Applications Symposium (SAS). DOI: 10.1109/SAS60918.2024.10636369} 
    \end{minipage}}}}%
  }
  \AtPageLowerLeft{%
    \raisebox{25pt}{\makebox[\paperwidth]{\begin{minipage}{21cm}\centering
      \textcolor{gray}{ \copyright 2024 IEEE.  Personal use of this material is permitted.  Permission from IEEE must be obtained for all other uses, in any current or future media, including reprinting/republishing this material for advertising or promotional purposes, creating new collective works, for resale or redistribution to servers or lists, or reuse of any copyrighted component of this work in other works.
      }
    \end{minipage}}}%
  }
}

\begin{document}

\title{Autonomous Navigation in Dynamic Human Environments with an Embedded 2D LiDAR-based Person Tracker}
\author{\IEEEauthorblockN{Davide Plozza\IEEEauthorrefmark{1},
       Steven~Marty\IEEEauthorrefmark{1},
       Cyril~Scherrer\IEEEauthorrefmark{1},
       Simon~Schwartz\IEEEauthorrefmark{1},
       Stefan~Zihlmann\IEEEauthorrefmark{1},
       Michele~Magno\IEEEauthorrefmark{1}}

       \IEEEauthorblockA{\IEEEauthorrefmark{1}Dept. of Information Technology and Electrical Engineering, ETH Z\"{u}rich, Switzerland}
       \IEEEauthorblockA{\{dplozza,martyste,cscherrer,sschwartz,szihlmann,magnom\}@ethz.ch}

       }

\newacronym{lidar}{LiDAR}{Light Detection and Ranging}
\newacronym{amr}{AMR}{Autonomous Mobile Robots}

\maketitle

\begin{abstract}
In the rapidly evolving landscape of autonomous mobile robots, the emphasis on seamless human-robot interactions has shifted towards autonomous decision-making.
This paper delves into the intricate challenges associated with robotic autonomy, focusing on navigation in dynamic environments shared with humans.
It introduces an embedded real-time tracking pipeline, integrated into a navigation planning framework for effective person tracking and avoidance, adapting a state-of-the-art 2D LiDAR-based human detection network and an efficient multi-object tracker.
By addressing the key components of detection, tracking, and planning separately, the proposed approach highlights the modularity and transferability of each component to other applications.
Our tracking approach is validated on a quadruped robot equipped with 270° 2D-LiDAR against motion capture system data, with the preferred configuration achieving an average MOTA of 85.45\% in three newly recorded datasets, while reliably running in real-time at 20 Hz on the NVIDIA Jetson Xavier NX embedded GPU-accelerated platform.
Furthermore, the integrated tracking and avoidance system is evaluated in real-world navigation experiments, demonstrating how accurate person tracking benefits the planner in optimizing the generated trajectories, enhancing its collision avoidance capabilities.
This paper contributes to safer human-robot cohabitation, blending recent advances in human detection with responsive planning to navigate shared spaces effectively and securely.

\end{abstract}


\section{Introduction}
\label{sec:intro}

As technology is evolving, the integration of \gls{amr} into our daily lives is accelerating, becoming an increasingly prevalent part of our everyday experiences. 
Commercially available robots offer numerous advantages, but one critical challenge remains: ensuring safety, particularly in environments where robots interact with humans~\cite{navarro2021proximity,rubagotti2022perceived}.
The autonomy in decision-making for a robot depends on comprehending its surroundings and its ability to react promptly to dynamic alterations in the environment~\cite{halme2018review}. For robots that operate close to people, it is the detection and tracking of humans within the robot's space. This important feature, not only guarantees the well-being of individuals but also maximizes the efficiency of the robots~\cite{lu2013towards}.

\gls{amr}, designed for a wide range of tasks from human assistance to industrial automation, frequently share the common objective of efficiently navigating to positional targets in a partially known environment.
Effective navigation hinges on the seamless integration of global path planning with local trajectory planning that ensures collision avoidance by dynamically responding to real-time sensor observations, including dynamic obstacles such as humans~\cite{pittner2018systematic}.
Accurately anticipating and understanding human trajectories is crucial for local trajectory planning algorithms, as this foresight allows the robot to proactively adjust its route to avoid potential collisions and optimize its motion in the presence of humans~\cite{medina2022perception}.

Various configurations using a range of sensors to detect and track humans in autonomous robots have been proposed in the recent literature, with the most popular sensors including RGB(-D) cameras~\cite{punn2021monitoring,xu2023onboard}, 2D LiDARs~\cite{jia2020dr,beyer2018deep,jung2013development}, and 3D LiDARs~\cite{3dlidarpersondetection}.
RGBD cameras provide richer environmental information but are limited by a narrower field of view and less precise distance measurements compared to LiDARs \cite{jia20222d}. 
This work focuses on 2D LiDARs over 3D due to their cost-effectiveness, energy efficiency, simpler data processing, and comparable accuracy for the person detection task\cite{HASAN2022393,jia20222d}.

Detection and tracking components are commonly validated using publicly available datasets~\cite{vendrow2023jrdb, martin2021jrdb}.
However, only a handful of studies have subjected the tracking pipeline to real-world applications, including tasks such as people-following by Leigh et al.\cite{JLT_tracker} or collision avoidance in this work. Such efforts are crucial, as comprehensive benchmarks assessing the system's overall performance are scarce but vital for ensuring safe and effective collision avoidance with moving humans~\cite{medina2022perception}.

To address these challenges, we present a real-time embedded 2D LiDAR-based person tracking pipeline integrated into a navigation planning framework for effective human avoidance. The pipeline merges a state-of-the-art human detector, optimized for a new sensor configuration, with an efficient multi-object tracker, achieving real-time performance on the NVIDIA Jetson Xavier NX embedded GPU-accelerated platform.
Experimental evaluations are conducted on a Unitree A1 quadrupedal robot to demonstrate and validate both tracking accuracy and human avoidance, utilizing three newly recorded datasets tailored for real-world use cases.

We summarize our contributions as follows:
\begin{itemize}
    \item Proposing an embedded 2D LiDAR-based person tracking pipeline, running at \qty{20}{Hz} on the Jetson Xavier NX.
    \item Integrating the tracker in a navigation planning framework.
    \item Conducting experimental evaluations with three newly collected real-world datasets.
\end{itemize}

\section{Related Works}
\label{sec:related}

Most recent works have moved from hand-crafted features to a data-driven approach for person detection using LiDAR.
Beyer et al. significantly improved person detection accuracy by incorporating CNNs in their novel approach called DROW~\cite{beyer2018deep}.
This work has been improved by DR-SPAAM, an auto-regressive neural network model with spatial attention by Jia et al.~\cite{jia2020dr}.
Due to great evaluation results on two datasets and fast enough interference speed for real-time usage~\cite{jia2021self}, we utilize it as the detection base for our pipeline.

Beyond DR-SPAAM, Jia et al. demonstrate the use of feature similarity between consecutive detections to establish tracklets for individual humans~\cite{jia2020dr}, but the integration of this detector in a multi-object tracking pipeline has yet to be explored, indicating the relevance of this work.
Existing 2D-Lidar-based human tracking works have relied on nearest neighbours association~\cite{jung2013development,JLT_tracker} coupled with Kalman Filters~\cite{chen2019pedestrian,jung2013development, JLT_tracker}.
Newer tracking methods like end-to-end approaches~\cite{wang2020towards,zeng2022motr} or Simple Online and Realtime Tracking (SORT)~\cite{sort_tracker} have yet to be explored for this purpose.

Tracks help predict human trajectories, enabling predictive planning for robot interactions, resulting in less disruption and enhanced safety compared to reactive planning~\cite{medina2022perception}. 
Different trajectory prediction models exist with varying assumptions and complexities~\cite{medina2022perception}. In our work, we primarily focus on collision avoidance and therefore employ a simple constant velocity model, as opposed to models that consider social norms~\cite{social_dyn_tracker} or predicting possible destination goals of humans~\cite{rosmann2017online}.
These trajectory predictions can be fed into a planning pipeline. 
One framework that integrates planning, and other aspects, is the Navigation Stack~\cite{ROS:navigation}.
Planning is typically split into a global path planner and a local trajectory planner~\cite{pittner2018systematic}.
The global planner exploits a map to determine the shortest path from an initial point to the destination with a search algorithm such as A*.
More crucial for collision avoidance is the local planner, which incorporates sensor information, such as human detections, to produce a collision-free trajectory that loosely follows the global path.
The most popular choice for this task is the optimization-based Timed-Elastic-Band (TEB)~\cite{teb_planner}.
TEB stands out for its reaction speed~\cite{pittner2018systematic} and ability to integrate trajectories of dynamic obstacles with fixed velocities into the planning~\cite{looi2021study}.
These attributes make it an ideal fit for a planner in a human-centric environment to predict possible collisions and avoid them as quickly as possible.

Human-aware robot navigation could also be achieved through specialized social planners with a focus on human-friendly and socially acceptable interactions~\cite{moller2021survey}.
Examples of this are HATEB~\cite{singamaneni2021human}, an adaptation of TEB designed for human awareness, or SA-CADRL, an end-to-end reinforcement learning-based approach~\cite{chen2017socially}.
As the objectives of these planners are less quantifiable compared to trajectory-based planners they are hard to validate~\cite {moller2021survey}.
This, along with the easier tuning and minimal computational overhead for real-time usage, motivates the use of trajectory-based planners like TEB.

Therefore our work not only contributes a state-of-the-art 2D LiDAR-based detection and tracking system but also highlights its real-world integration within a local planning framework, addressing essential aspects often overlooked in current literature.

\section{Methods}
\label{sec:methods}

As explained in the related work from \cref{sec:related}, a successful people avoidance pipeline requires three key components: detection, tracking, and local planning.
Instead of an end-to-end approach, this paper addresses these three problems separately.
This modular approach allows for the independent transferability of each component to other applications.
While each component is individually fine-tuned and benchmarked, they are integrated to address challenges arising from the preceding steps.

\subsection{System}\label{subsec:system}
\begin{figure}
\centering
\includegraphics[width=0.8\columnwidth]{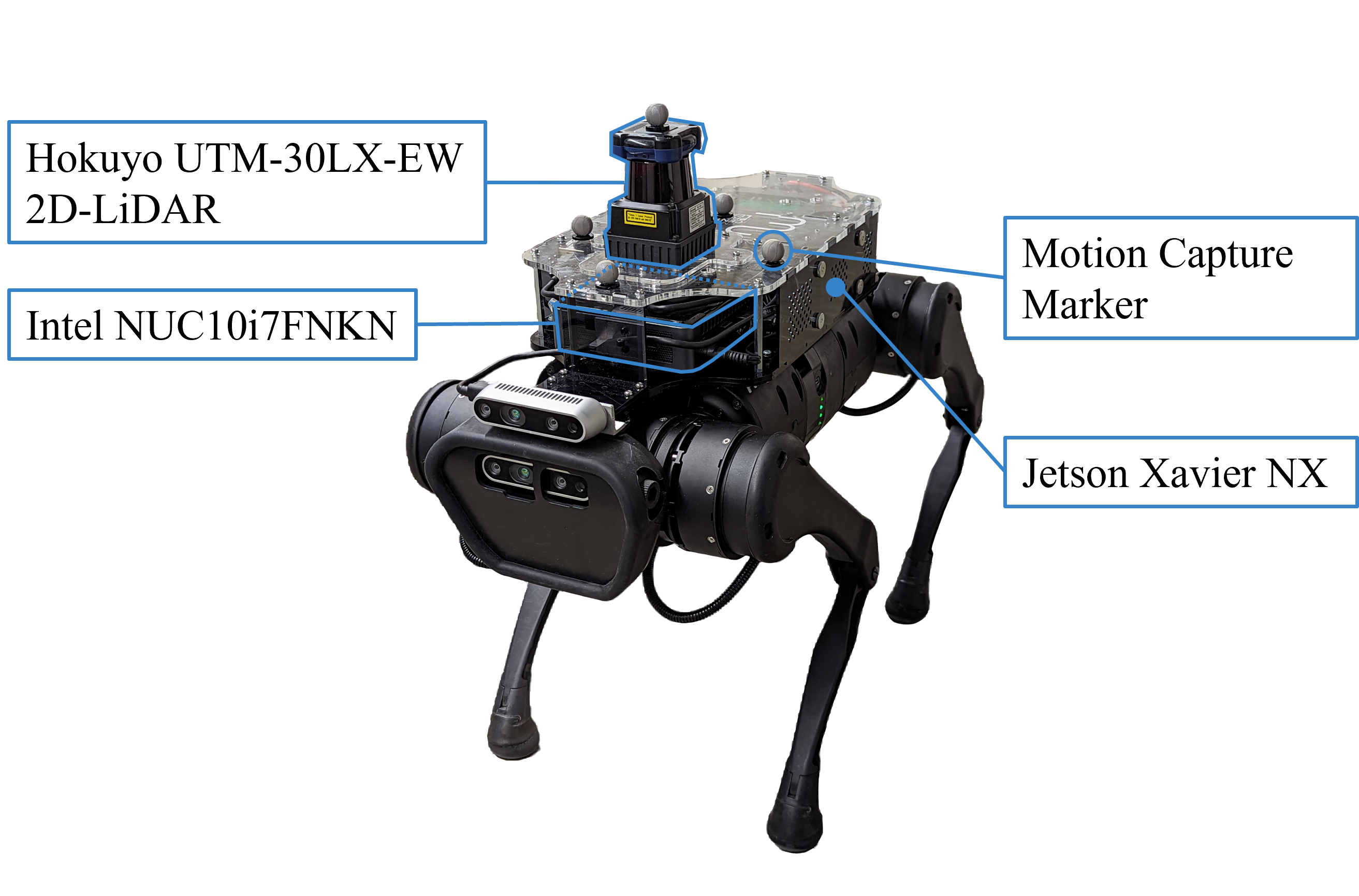}
\caption{Unitree A1 robot used in this work with the additional backpack system.
}
\label{fig:robodog}
\end{figure}

In this study, we employ a Hokuyo UTM-30LX-EW 2D LiDAR configured with \qty{20}{Hz} scan rate, a 270° scan angle and 0.25° angular resolution, yielding 1080 points per scan. The LiDAR is mounted horizontally at roughly \qty{45}{cm} from the ground.
The processing is carried out on an Intel NUC10i7FNKN with 10th gen 6-core i7 CPU, alongside a NVIDIA Jetson Xavier NX.
As seen in \cref{fig:robodog}, these components are mounted on the A1 legged robot from Unitree, inside a custom-built backpack, capable of integrating additional computing units, batteries, or sensors.
Velocity commands are translated into motor commands using Unitree's official walking controller, which also provides odometry data at \qty{50}{Hz}, and the avoidance pipeline is developed using ROS~\cite{ros_citation}.

\subsection{Detection}
The implementation of the novel 2D-LiDAR human detection models DR-SPAAM~\cite{jia2020dr}, mentioned in \cref{sec:related}, is open-source~\cite{GIT:drow_drspaam} and therefore integrated in this work.
DR-SPAAM utilizes a 1D convolutional neural network on a one-meter window around each LIDAR point to predict people's 2D locations relative to the LiDAR frame, along with a confidence value.
A voting scheme combines these predictions.
The fixed one-meter window is resampled to account for different LiDAR resolutions at various distances to ensure a consistent sample count around each point.
This design, proven by the CROWDBOT project~\cite{web:crowdbot}, allows successful deployment on diverse robots and sensors without retraining for the specific sensor, demonstrating angular resolution and distance independence.
Consequently, we opted not to create a new dataset with our LiDAR but instead chose to utilize the JRDB~\cite{martin2021jrdb}, which offers a 360° field of view with a 0.32° angular resolution, similar to our 0.25°

It's important to note that these networks are deployed on the Jetson Xavier NX platform, distinct from their original evaluation settings, which may affect reported interference speeds.
Notably, interference speed has a critical role in the tracker and avoidance components, given that the detector constrains the processing speed of the subsequent pipeline as seen in \cref{fig:pipelining}.
The inference time of one LiDAR frame highly depends on the number of windows processed by the network. 
While DR-SPAAM authors primarily report the effects of frame stride, which involves skipping full frames, their implementation also offers a window stride, essentially skipping some LiDAR points and the corresponding windows.
Our configurations presented in \cref{tab:configs} adapt this setting.

\subsection{Tracking}
\label{subsec:tracking}

Efficient real-time multi-object tracking is critical for our application, thus we integrate the open-source Norfair tracker \cite{norfair_tracker} in our pipeline. It is a lightweight Python library optimized for real-time video processing applications, maintained by Tryolabs.
The tracker adapts the SORT algorithm \cite{sort_tracker}, generalizing the point tracking to a variable number of points per detection.

Our contribution lies in tailoring Norfair for the people tracking task from LiDAR-based person detections. 
While it is optimized to work with computer vision-based detectors, Norfair's versatility allows it to be detector-agnostic, accepting two-dimensional inputs, making it an excellent candidate for our tracking needs.

The tracker uses a Kalman filter for motion estimation, with a constant velocity model.
The data association is efficiently solved using the Hungarian algorithm, which we configure with an euclidean cost function.
While sophisticated probabilistic data association methods could offer improved results, they are not practical for real-time online tracking due to their complexity~\cite{sort_tracker}.

Person track initiation and deletion are based on the count of matches to detections: each track keeps a counter, incremented during the tracker's update if for a match or decreased otherwise.
A candidate track is created from each unmatched detection, but it is only properly initiated after the counter reaches \( C_{init} \).
Existing tracks are terminated if there are no matches for over \( C_{del} \) updates. 
For effective obstacle avoidance, it is important to minimize the track initiation time, which is the duration from a subject's first detectable appearance to the proper initiation of its track. 
Balancing this with false positives requires careful tuning of \( C_{init} \) and the detector confidence threshold.
By relying on a robust detector, we can set the confidence threshold fairly high ($\geq0.8$) to minimize false positive detections.

To compensate for the robot's motion, we transform the detections into the robot's odometry frame of reference, where the tracking is then performed, which is a critical step for accurate motion estimation.

\begin{figure}
\centerline{\includegraphics[width=\columnwidth]{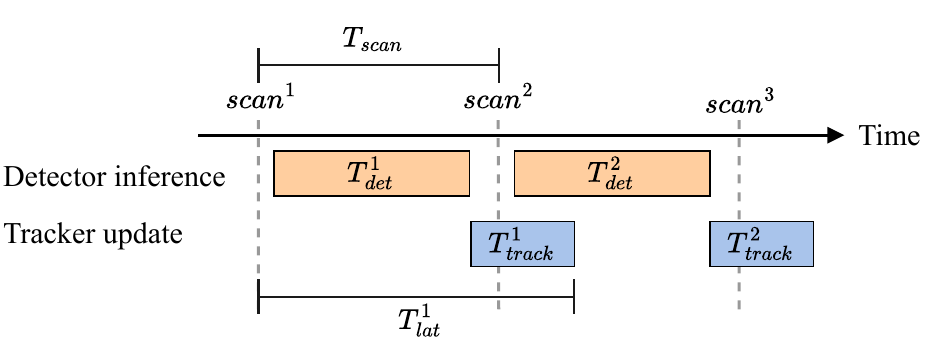}}
\caption{Pipelined execution of detector (inference time \(T_{det}^{i}\)) and tracker (update time \(T_{track}^{i}\)) for each LiDAR scan \(i\), received regularly with period \(T_{scan}\). The overall detection and tracking pipeline update latency is \(T_{lat}^{i}\).
}
\label{fig:pipelining}
\end{figure}

The detector inference and the tracker update are performed sequentially after each LiDAR scan is received. The execution is pipelined, as shown in \cref{fig:pipelining}, enabling operating at higher frequencies, which has a positive effect on both the inference as well as tracker accuracies.
The tracker is deployed alongside the detector on the Jetson Xavier NX platform.

\subsection{Planning and Navigation}
The planning comprises key components: mapping, localization, global planning, and local planning. We use the ROS Navigation Stack~\cite{ROS:navigation}, configured AMCL for localization and A* as global planner.
Our focus lies on the local planner, namely TEB.
TEB optimizes the robot's trajectory, considering variables like speed, time constraints, and environmental factors, ensuring efficient and adaptable navigation.
In addition to reacting to static obstacles in the map around the robot, dynamic obstacles with linear velocities can be introduced into the optimization process.
These dynamic obstacles and their velocities are estimated by the tracker from~\cref{subsec:tracking}.
Obstacles with a smaller velocity magnitude than \qty{0.05}{m/s} are ignored, a threshold sufficient to filter out most false positives.
The maximum speed of the robot is set to \qty{0.5}{m/s}.
One problem when incorporating dynamic obstacles into the planning process is that TEB also identifies them as static obstacles, preventing the robot from executing efficient avoidance measures.
To address this issue, the static obstacles around the tracked person's location are filtered out.

\section{Experimental Results}

\begin{table}
\centering
\caption{Configurations}
\label{tab:configs}
\begin{tabular}{lllll}
\toprule
Config                             & Window Stride & Conf. Thresh. & $C_{init}$ & $C_{del}$ \\
\midrule
\textit{Config-1} & 1      & 0.85         & 10    & 15   \\
\textit{Config-2} & 10     & 0.85         & 10    & 15   \\
\textit{Config-3} & 10     & 0.8          & 5     & 15  \\
\bottomrule
\end{tabular}
\end{table}

\label{sec:results}

\begin{table}
\centering
\caption{Benchmark Results.}
\label{tab:vicon_bench}
\begin{tabular}{@{}L{0.6cm}L{1.0cm}L{0.75cm}L{0.7cm}L{0.6cm}L{0.6cm}L{0.9cm}L{0.9cm}@{}}
\toprule
DS         & Tracker           & Valid & ID Switch & Miss & FP & MOTA &MOTP [m] \\ \midrule
  \multirow{3}{*}{SR}
      & \textit{Config-1}  & 7421   & \textbf{9}       & 424 & \textbf{7}     & 94.40\%  & 0.13      \\
      & \textit{Config-2}  & 7431   & 11      & 412 & 12    & \textbf{94.46}\%  & 0.13     \\
      & \textit{Config-3}  & \textbf{7523}   & 10      & \textbf{321} & 112   & 94.26\%  & 0.13    \\
      
\midrule
  \multirow{2}{*}{MR1}
      & \textit{Config-1}  & 9062    & 11    & 593 & \textbf{291}    & \textbf{90.74}\%  & 0.17     \\
      & \textit{Config-2}  & 9240    & 7     & 419 & 485    & 90.58\%  & 0.17    \\
      & \textit{Config-3}  & \textbf{9443}    & \textbf{6}     & \textbf{217} & 1607   & 81.07\% & 0.17    \\
\midrule
\multirow{2}{*}{MR2}
      & \textit{Config-1}  & 5667    & 47    & 1052  & 338    & 78.76\% & 0.18   \\
      & \textit{Config-2}  & 5789    & \textbf{45}    & 932   & \textbf{292}    & \textbf{81.26}\% & 0.18   \\
      & \textit{Config-3}  & \textbf{6243}    & 46    & \textbf{477}   & 762    & 81.01\% & 0.18    \\
\bottomrule
\end{tabular}
\end{table}

Tracking accuracy and human avoidance are evaluated with a custom real-world dataset and integrated experiments.
Three configurations of our tracker, specified in \cref{tab:configs}, are evaluated: \textit{Config-1} optimizes overall tracking accuracy, \textit{Config-2}  targets real-time usage and \textit{Config-3} reduces track initiation time.

\subsection{Datasets}

Three new datasets are recorded to benchmark the performance of our tracking system, tailored to the context of obstacle avoidance.
We use a motion capture system based on six Vicon Vero 2.2 cameras to record the ground truth position and orientation of both the robot and the people at \qty{100}{Hz}. Sensor and odometry data is recorded from the robot configuration described in~\cref{subsec:system}.

In each experiment, three participants walk in straight lines for the first half, and then randomly for the second half, simulating various pedestrian behaviours. Both the people and the robot are confined in a \qty{4}{\meter}x\qty{4}{\meter}  area inside one room with moderate clutter at the edges in the form of chairs and tables.
This scenario is ideal for testing the system's tracking performance in close-range scenarios critical for obstacle avoidance.

The datasets are as follows:
\begin{enumerate}
    \item \emph{Stationary Robot (SR)}: The robot remains stationary, providing a controlled baseline for performance evaluation. Duration: 2 minutes and 11 seconds.
    \item \emph{Moving Robot 1 (MR1)}: The robot moves semi-randomly at up to \qty{0.5}{\meter\per\second} (forward/backward) and \qty{\pm 1.5}{\radian\per\second} (rotation), ensuring participants are inside the LiDAR field of view at any time. Duration: 2 minutes and 41 seconds.
    \item \emph{Moving Robot 2 (MR2)}: The robot moves randomly with the same speed limits as \emph{MR1} with participants frequently leaving and re-entering the LiDAR's field of view. Duration: 2 minutes and 24 seconds.
\end{enumerate}

\begin{figure}[]
\centering
\includegraphics[width=\columnwidth]{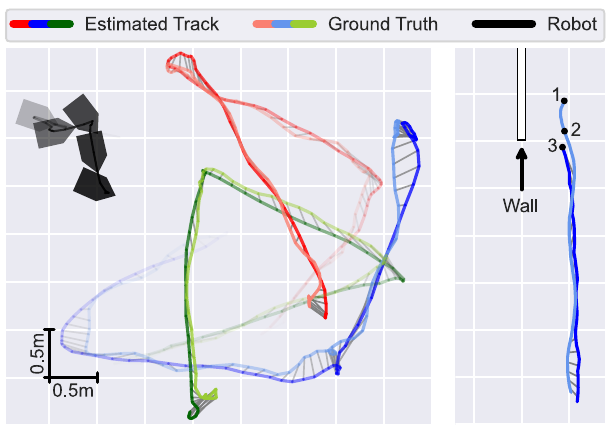}
\begin{minipage}[t]{.72\linewidth}

\centering
\vspace{-0.5cm} 
\subcaption{}\label{fig:trajectories}
\end{minipage}%
\begin{minipage}[t]{.28\linewidth}
\centering
\vspace{-0.5cm} 
\subcaption{}\label{fig:latency}
\end{minipage}
\captionsetup{subrefformat=parens}

\vspace{-0.1cm} 
\caption{Trajectories of the tracker with \emph{Config-3} together with the ground truth from the Vicon Motion Capture system.\\ Subfigure~\subref{fig:trajectories} depicts the tracker's performance with three randomly moving individuals, displaying time using transparency.\\ Subfigure~\subref{fig:latency} displays the track initiation time of a person emerging from behind a wall, where the robot is on the left side of the wall (1: starting point person, 2: first observation of person with at least five LIDAR points, 3: track initiation).}
\label{fig:tracks}
\end{figure}

\cref{fig:trajectories} shows a qualitative visualization of the ground truth and predicted trajectories on a short section of the \emph{MR1} dataset.

\subsection{Benchmark}

The CLEAR MOT metrics \cite{clear_mot} are utilized for the quantitative analysis of our detection and tracking pipeline.
The evaluation is based on the open-source benchmark framework proposed by~\cite{JLT_tracker}, specifically the Multi-Object Tracking Accuracy (MOTA) and the Multi-Object Tracking Precision (MOTP).
The MOTA score, detailed in Equation~(\ref{eq:mota}), quantifies overall accuracy by accounting for identity switches (\( \text{ID}_k \)), misses (\( \text{Miss}_k \)), and false positives (\( \text{FP}_k \)), normalized by the number of ground truths (\( g_k \)). Since the MOTA score alone does not differentiate the impact of the different error types, individual error counts are also reported. Ground truth annotations and track estimations are matched using a threshold of \qty{0.75}{m}.

\begin{equation}
MOTA = 1 - \frac{\sum_k (ID_k + Miss_k + FP_k)}{\sum_k g_k} \label{eq:mota}
\end{equation}

MOTP, shown in Equation~(\ref{eq:motp}), provides a measure of the tracker's precision, by averaging the distances (\( d_{k}^{i} \)) between matched estimates and ground truths, over all correct matches (\( c_k \)). 
A lower MOTP indicates higher precision.

\begin{equation}
MOTP = \frac{\sum_{i,k} d_{k}^{i}}{\sum_k c_k} \label{eq:motp}
\end{equation}

Ground truth annotations are systematically extracted from the motion capture data regardless of whether the person is visible by the LiDAR. Then at each timestamp, only the annotations and estimations inside the LiDAR field of view were considered. This approach still ensures a fair benchmark.

Table~\ref{tab:vicon_bench} shows the benchmark results, with CLEAR MOT metrics and the raw counts of each error type.
In the \emph{SR} dataset, all configurations of our achieve a very high MOTA score.
The performance of our tracker remains high in the more dynamic setting of the \emph{MR1} dataset, especially for \emph{Config-1} and \emph{Config-2}, illustrating robustness in semi-random movement scenarios.
In the challenging \emph{MR2} dataset all configurations show a drop in MOTA score, with more ID switches and misses caused by the fact that subjects frequently exit the LiDAR's field of view. \emph{Config-1} and \emph{Config-2} have similar scores across all datasets, showing a small effect of the stride parameter.
All configurations show similar MOTP, which stays below \qty{0.2}{m/s} even in the most challenging configuration.

\emph{Config-3} as expected trades fewer misses for a higher count of false positives, while having an average MOTA of 85.45\% across all datasets, in contrast to \emph{Config-2}'s 89.99\%.
The low track initiation time of \emph{Config-3} can be seen in an additional experiment shown in \cref{fig:latency}, as a person emerges from behind a barrier at approximately \qty{1}{m/s}. 
The proximity of points 2 and 3 demonstrates \emph{Config-3}'s prompt initiation of tracking upon the person's first detectable appearance.

\subsection{Computation Time}

The computation times of each configuration of our tracker are benchmarked on a Jetson Xavier NX. We report the detector inference time \(T_{det}^{i}\), the tracker update time \(T_{track}^{i}\), and the total update latency time \(T_{lat}^{i}\), all described in  \cref{fig:pipelining}. The average and worst times are measured across all three datasets \emph{SR}, \emph{MR1} and \emph{MR2}.
While the \emph{Config-1} is not fast enough for real-time execution at \qty{20}{Hz}, both \emph{Config-2} and  \emph{Config-3} maintain an average processing time well under the scan period \(T_{det}^{i}\) of \qty{50}{ms}.
Importantly, while the total worst times exceed the scan period, both \(T_{det}^{i}\) and \(T_{track}^{i}\) are always below \qty{50}{ms} even in the worst-case scenario, highlighting the effectiveness of the pipelined execution.

\begin{table}[]
\centering
\caption{Computation Times.}
\label{tab:runtime}
\begin{tabular}{@{}lL{2.1cm}L{2.1cm}L{2.1cm}@{}}
\toprule
Tracker     & Detector \(T_{det}^{i}\) Worst/Avg [ms] & Tracker \(T_{track}^{i}\)  Worst/Avg [ms] & Total time \(T_{lat}^{i}\) Worst/Avg [ms]  \\ \midrule
\emph{Config-1}      &  196.76/172.83    &  44.96/14.11  &  233.37/186.94    \\
\emph{Config-2}      &  \textbf{43.04}/31.62      &  27.42/\textbf{7.66 }  &  64.93/\textbf{39.28}   \\
\emph{Config-3}      &  43.52/\textbf{31.54 }     &  \textbf{25.51}/7.99   &  \textbf{61.31}/39.53   \\ \bottomrule
\end{tabular}
\end{table}

\begin{figure}[]
\centering
\includegraphics[width=\columnwidth]{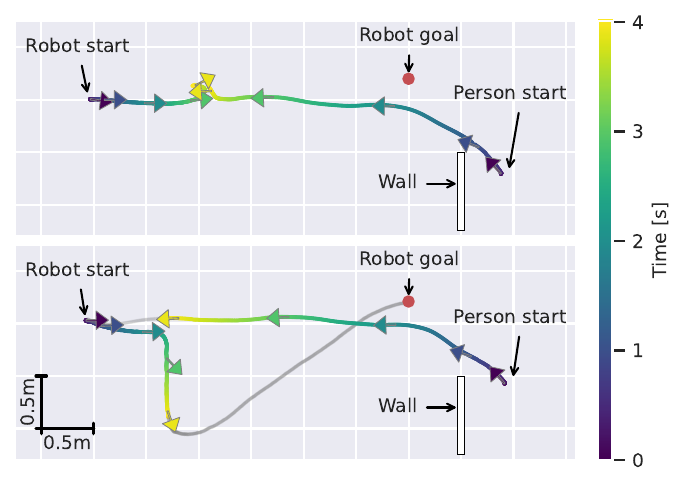}

\vspace{-0.1cm} 
\caption{Comparative navigation experiments illustrating the robot and human trajectories from the Vicon Motion Capture system. TEB is configured without our tracker (top) where it fails to avoid collision, and with \emph{Config-3} (bottom) where it successfully performs an early avoidance maneuver.}
\label{fig:vicon_teb_exp}
\end{figure}

\subsection{Navigation experiments}

Finally, the whole tracking pipeline together with the TEB local planner is evaluated in the avoidance task.
This test assesses the robot's ability to avoid a frontal collision with a person walking towards it as seen in \cref{fig:vicon_teb_exp}.
To demonstrate the improved avoidance performance, the test is conducted both without our tracker, treating all detected obstacles as static, and with \emph{Config-3} of our pipeline, exploiting the prediction of the human trajectory.
In both experiments, designed to test both the tracker and planner reaction time, the robot is instructed to reach a position in front of him, while the person starts walking from behind a wall towards the robot at approximately \qty{1}{m/s} while being initially occluded from the robot's LiDAR.
This is a challenging situation for obstacle avoidance, as the robot has little time to react due to its slow maximum speed of \qty{0.5}{m/s}.
\cref{fig:vicon_teb_exp} shows that without our tracker the robot reacts too late to avoid a collision, while with the help of our pipeline, the robot performs an avoidance maneuver soon enough.

\section{Discussion}

Our results, based on a new LIDAR setup without retraining the network, confirm the DR-SPAAM's resolution and distance independence~\cite{jia2020dr}, achieving high detection accuracy for successful human tracking. 
The benchmark on our custom dataset shows that all configurations of our proposed tracking pipeline achieve robust close-range human tracking in partially cluttered environments, even on the challenging \emph{MR2} dataset where ID switches are bound to be higher. 
\emph{Config-3} demonstrates a low track initiation time and number of misses, crucial for timely detection and reaction to dynamic obstacles, without sacrificing too much on the MOTA score. These attributes establish it as our favored configuration for people avoidance applications.
The similar performance of \emph{Config-1} and \emph{Config-2} shows that the stride parameter has a small effect on close-range detections.
The influence of this parameter on larger-range scenarios must still be evaluated, but even if it worsens those results, they are less critical compared to the avoidance and planning occurring at close range.
Our approach shows consistent computation time, with \emph{Config-2} and \emph{Config-3} reliably running in real-time at \qty{20}{Hz} on the Jetson Xavier NX.

The navigation experiments demonstrate that our pipeline, coupled with TEB, significantly enhances collision avoidance, especially in avoiding people in collision routes with the robot at normal walking speeds. Our approach consistently keeps a safe minimum distance from the person.
The improved planning efficiency with our tracker could be further explored with additional experiments in different scenarios.
The simple linear motion model may struggle with unexpected human reactions, warranting exploration of more complex models for such cases.

We also recognize limitations such as data assignment challenges during temporary occlusions and a high number of false positives while the robot is in motion. The latter problem could be mitigated by integrating a local map from the navigation stack, to filter our false positives near walls.

\section{Conclusion}

This paper proposes a fully embedded real-time 2D LiDAR-based person-tracking pipeline for enhancing obstacle avoidance in dynamic human environments.
By integrating the pipeline in an autonomous navigation planning framework, and deploying the system on a quadrupedal robot, we show that our approach not only achieves consistent real-time performance at \qty{20}{Hz} but also effectively enables the robot to safely navigate around humans.
The high MOTA score of 85.45\% achieved by \emph{Config-3} in the tracking benchmark on our custom datasets validates the performance of our pipeline in dynamic environments.
This work lays a foundational step towards the integration of autonomous robots into human-centric environments, offering significant implications for applications such as navigation assistance for individuals with vision impairments.

Future work will focus on further optimizing the pipeline by fine-tuning both the detector and tracker, benchmarking the tracking performance against other approaches, such as \cite{JLT_tracker}, and performing a more extensive real-world evaluation of the avoidance capabilities to quantify the safety of such system.

\section*{Acknowledgment}

This work was supported by the ETH Future Computing Laboratory (EFCL).

\bibliographystyle{IEEEtran}
\bibliography{IEEEabrv,bibtex/bib/citations}

\end{document}